\documentclass[lettersize,journal]{IEEEtran}
\usepackage{amsmath,amsfonts}
\usepackage{algorithmic}
\usepackage{algorithm}
\usepackage{array}
\usepackage[caption=false,font=normalsize,labelfont=sf,textfont=sf]{subfig}
\usepackage{textcomp}
\usepackage{stfloats}
\usepackage{url}
\usepackage{verbatim}
\usepackage{graphicx}
\usepackage{cite}
\usepackage{graphicx, multirow, booktabs, ctable, pifont}
\usepackage{hyperref}
\usepackage{balance}
\usepackage[symbol]{footmisc}

\usepackage{soul,color}
\soulregister\cite7
\soulregister\ref7

\usepackage{xcolor}
\usepackage{amssymb}

\begin{document}

\title{Confidence-Aware Paced-Curriculum Learning by Label Smoothing
for Surgical Scene Understanding}




\author{Mengya Xu$^{1,2,3\dagger}$, Mobarakol Islam$^{4\dagger}$, Ben Glocker$^{4}$ and Hongliang Ren$^{1,2,3}$

\thanks{This work was supported by the National Key R\&D Program of China under Grant 2018YFB1307700 (with subprogram 2018YFB1307703) from the Ministry of Science and Technology (MOST) of China, Hong Kong Research Grants Council (RGC) Collaborative Research Fund (CRF C4026-21GF), the Shun Hing Institute of Advanced Engineering (SHIAE project BME-p1-21, 8115064) at the Chinese University of Hong Kong (CUHK), and Singapore Academic Research Fund under Grant R397000353114. This work also has the support of Wang An. Thank him for participating in the discussion of the project.} 

\thanks{$^{\dagger}$Mengya Xu and Mobarakol Islam are co-first authors.}%
\thanks{$^{1}$Mengya Xu and Hongliang Ren are with Dept. of Biomedical Engineering, National University of Singapore, Singapore; Dept. of Electronic Engineering, The Chinese University of Hong Kong, Hong Kong SAR, China; National University of Singapore (Suzhou) Research Institute (NUSRI), China (\href{mailto:mengya@u.nus.edu}{mengya@u.nus.edu}, \href{mailto:hlren@ieee.org}{hlren@ieee.org})}
\thanks{$^{2}$Mengya Xu and Hongliang Ren are with Dept. of Electronic Engineering, The Chinese University of Hong Kong, Hong Kong SAR, China}
\thanks{$^{3}$Mengya Xu and Hongliang Ren are with National University of Singapore (Suzhou) Research Institute (NUSRI), China}
\thanks{$^{4}$Mobarakol Islam and Ben Glocker are with BioMedIA Group, Imperial College London, UK. (\href{mailto:m.islam20@imperial.ac.uk}{m.islam20@imperial.ac.uk}, \href{mailto:b.glocker@imperial.ac.uk}{b.glocker@imperial.ac.uk})}
}

\maketitle

\begin{abstract}
Curriculum learning and self-paced learning are the training strategies that gradually feed the samples from easy to more complex. They have captivated increasing attention due to their excellent performance in robotic vision. Most recent works focus on designing curricula based on difficulty levels in input samples or smoothing the feature maps. However, smoothing labels to control the learning utility in a curriculum manner is still unexplored. In this work, we design a paced curriculum by label smoothing (P-CBLS) using paced learning with uniform label smoothing (ULS) for classification tasks and fuse uniform and spatially varying label smoothing (SVLS) for semantic segmentation tasks in a curriculum manner. In ULS and SVLS, a bigger smoothing factor value enforces a heavy smoothing penalty in the true label and limits learning less information. Therefore, we design the curriculum by label smoothing (CBLS). We set a bigger smoothing value at the beginning of training and gradually decreased it to zero to control the model learning utility from lower to higher. We also designed a confidence-aware pacing function and combined it with our CBLS to investigate the benefits of various curricula. The proposed techniques are validated on four robotic surgery datasets of multi-class, multi-label classification, captioning, and segmentation tasks. We also investigate the robustness of our method by corrupting validation data into different severity levels. Our extensive analysis shows that the proposed method improves prediction accuracy and robustness. The code is publicly available at \url{https://github.com/XuMengyaAmy/P-CBLS}.

\emph{Note to Practitioners}--The motivation of this article is to improve the performance and robustness of deep neural networks in safety-critical applications such as robotic surgery by controlling the learning ability of the model in a curriculum learning manner and allowing the model to imitate the cognitive process of humans and animals. The designed approaches do not add parameters that require additional computational resources.

\end{abstract}



\section{Introduction}
\IEEEPARstart{S}{urgical} scene understanding ability acquired with the help of deep neural networks (DNNs) is essential for developing ambient clinical intelligence. It allows for intraoperative assistance and postoperative analysis and ensures effective treatment. Despite the high performance of deep neural networks (DNNs), poor generalization, less robustness, and miscalibration issues have limited their use in safety-critical applications such as robotic surgery, medical diagnosis, and autonomous driving. The model performance often degrades with the distribution shift arising from images such as domain shift, population shift, and acquisition shift~\cite{rabanser2018failing, castro2020causality, stacke2020measuring}. There are also studies to present the miscalibration of the DNNs where model prediction is overconfident and less trustworthy~\cite{guo2017calibration, mukhoti2020calibrating}. These lead to an emphasis on designing a more robust and generalized model to provide safe and reliable predictions in sensitive applications.

\begin{figure*}[!hbpt]
\centering
\includegraphics[width=1\linewidth]{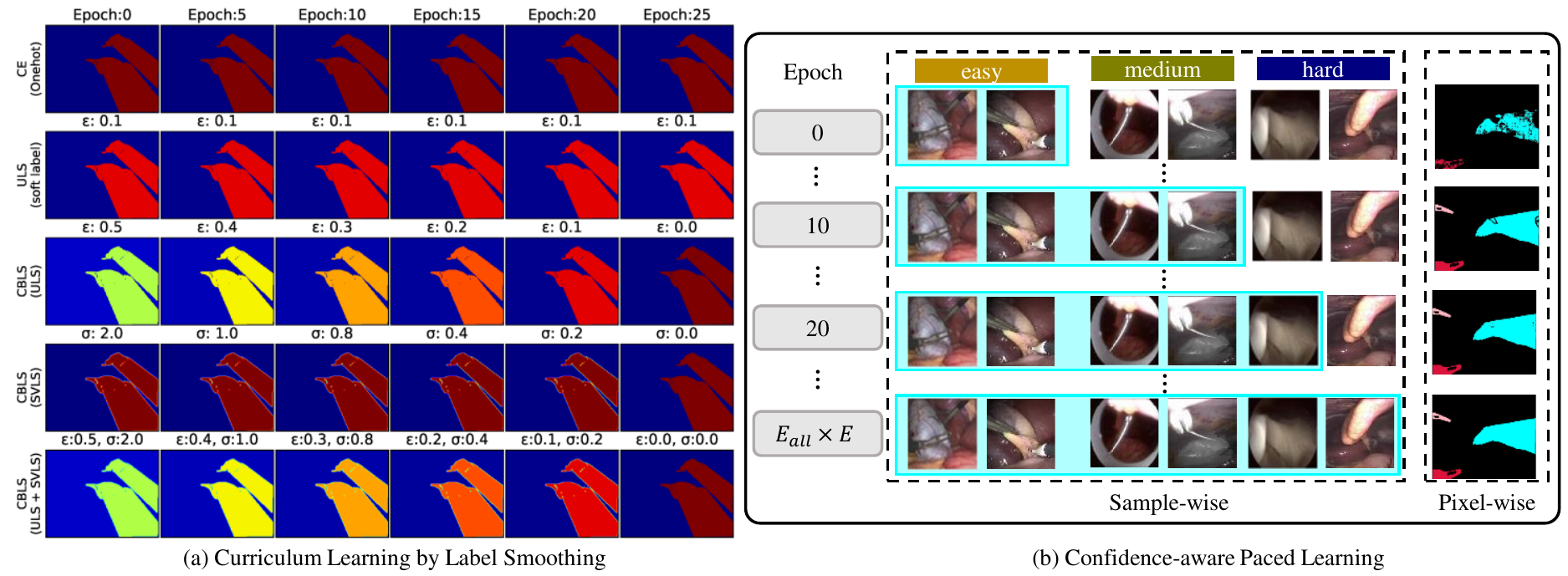}
\caption{\textbf{Confidence-Aware Paced-Curriculum Learning by
Label Smoothing} (a) Curriculum Learning by Label Smoothing. one-hot for cross-entropy (CE) and smoothed labels for uniform label smoothing (ULS) and various CBLS versions over the training epochs. The different colors of the instrument area displayed in the label map reflect different smoothing factor values. CE (one-hot) sets the smoothing factor $\epsilon$ to $0$. Thus the instrument area is always dark red over the training epochs. ULS (soft label) set $\epsilon$ to $1$. The instrument area keeps rose red with training. Our CBLS contains three different versions: ULS, SVLS, and ULS+SVLS. As training progresses, we decrease the smoothing factor of ULS ($\epsilon$), SVLS ($\sigma$), and ULS+SVLS ($\epsilon$ and $\sigma$) gradually to squeeze the label in a curriculum way that changes the true probability value from lower to higher. In the label map of CBLS (ULS), the instruments area changes from green to dark red as the $\epsilon$ decreases. The color of the instrument contour is the same as the interior color. In the label map of CBLS (SVLS), the smoothing factor $\epsilon$ of ULS remains at $0$, and the smoothing factor $\sigma$ of SVLS decreases gradually. $\sigma$ mainly affects the value at the contour. Therefore, the color of the instrument contour is different from the interior color. Meanwhile, the interior color remains the same over the training epochs. In the label map of CBLS (ULS+SVLS), the color at the instrument contour is also different from the interior color. However, the interior color is also gradually changing as the smoothing factor $\epsilon$ of ULS decreases. (b) Confidence-aware Paced Learning, including sample-wise paced learning, and pixel-wise paced learning. The samples or pixels are chosen in the order of easier to harder levels based on their confidence score during training.}
\label{fig:main}
\end{figure*}

Recently, curriculum learning~\cite{bengio2009curriculum} and self-paced learning~\cite{kumar2010self} have attracted the interest of people in machine learning and computer vision due to their outstanding generalization ability. Both learning paradigms are based on the learning principle that underpins the cognitive process of humans and animals, ordering samples based on their difficulty levels during the model training. Several studies have observed that curriculum learning can significantly improve the generalization and convergence speed in computer vision~\cite{li2021curriculum, wang2019dynamic, guo2018curriculumnet, li2017self} and natural language processing~\cite{platanios2019competence}. The performance of curriculum learning mostly depends on how accurate is the difficulty measurement technique. Most previous works have used confidence-based difficulty scores from a baseline model~\cite{zhao2021knowledge, panagiotatos2019curriculum}. Handcrafted features~\cite{zhu2021combining}, annotator agreement~\cite{wei2021learn}, and similarity scores~\cite{zhang2019curriculum} are also utilized to sort the samples from more uncomplicated to more complex. Most recently, Sinha et al.~\cite{sinha2020curriculum} apply Gaussian smoothing from higher to lower gradually on the feature maps to control information flow during training. However, all these works either focus on input samples or feature maps to design the curriculum scheme without considering the adjustment of learning utility through label smoothing. 

In this work, we explore label smoothing~\cite{szegedy2016rethinking}, a regularization method that can penalize the over-confidence prediction~\cite{muller2019does} by flattening the hard targets into the soft labels, to design Curriculum By Label Smoothing (CBLS). We utilize three variations of label smoothing, (i) uniform label smoothing (ULS)~\cite{szegedy2016rethinking}, (ii) spatially varying label smoothing(SVLS)~\cite{islam2021spatially}, and (iii) fused ULS and SVLS to develop our CBLS. We vary the smoothing factor from higher to lower over the training epochs to flatten the hard target (one-hot) to control learning utility in a curriculum manner. Fig.~\ref{fig:main}(a) demonstrates the hard label and soft labels for different learning strategies such as cross-entropy (CE), ULS, proposed CBLS (ULS), CBLS (SVLS), and CBLS (ULS + SVLS). We also design a confidence-aware pacing function (see Fig.~\ref{fig:main}(b)) to conduct extensive analysis between different types of curriculum techniques and their individual and combined benefits. Our main contributions are summarized as follows:

\begin{itemize}
\item We design a novel curriculum learning strategy, Curriculum By Label Smoothing (CBLS), by smoothing labels in a curriculum manner that controls the learning utility from lower to higher;

\item We develop a confidence-aware pacing function to order from easy to more complex samples for classification tasks and pixels for segmentation tasks and build paced-CBLS (P-CBLS) to investigate the benefits of both curricula;

\item Our method shows improved performance over multiple baselines for four robot vision recognition tasks of multi-class, multi-label classification, captioning, and segmentation;

\item We also investigate the robustness of CBLS by corrupting images with different severity levels, and the results suggest the constantly improved prediction with the severity increases.  

\end{itemize}

\section{Related Work}

\subsection{Curriculum learning / self-paced learning} Inspired by human and animal learning principles, curriculum learning introduces samples from easier to the complex  during training. It is observed that the learning process from easy to difficult tasks helps achieve better performance by avoiding the local minima and obtaining better generalization results~\cite{basu2013teaching, tang2012self}. Curriculum learning and self-paced learning also improve the robustness and reliability of noisy samples~\cite{li2017self, wang2019dynamic}. There are several ways to measure the sample difficulty in designing this learning technique. Most of the works adopt the confidence score to sort the samples. In this way, first, feed the high confidence/easy samples and subsequently introduce low confidence/difficult ones into the learning. Previous studies also utilized handcrafted features~\cite{zhu2021combining}, multi-raters disagreement~\cite{wei2021learn} and similarity scores~\cite{zhang2019curriculum}. A cutting-edge strategy~\cite{kepple2021curriculum} is introduced to use curricula to pinpoint the fundamentals of how a system learns. Domain-aware Curriculum Learning~\cite{roy2021curriculum} identifies curriculum learning as one crucial element that can reduce the multiple domain shifts in the multi-target domain adaptation. It adapts to the easier target domains first, then moves on to the more difficult ones. A curriculum based on human visual acuity~\cite{basu2022surpassing} lessens the texture biases in models for gallbladder cancer. Most recently, the Curriculum By Smoothing (CBS)~\cite{sinha2020curriculumwH} employs the Gaussian filter of feature maps from a higher variance to lower across the training epochs. The higher variance smoothes the feature map heavily and limits the model to learn less information at the beginning. However, designing a curriculum by smoothing label probability is still unexplored in this domain.

\subsection{Label smoothing}
Label smoothing is originally proposed by~\cite{szegedy2016rethinking} as a learning strategy to improve the prediction. We name it uniform label smoothing (ULS) in our work. Many classification models~\cite{zoph2018learning, real2019regularized, huang2019efficient} incorporate ULS as the regularization technique to improve the model learning. Most recently, ULS has been found to be a calibration technique that limits the overly-confidence prediction by flattening full probability in the hard targets~\cite{muller2019does}. There is also evidence that ULS can improve the feature representation and boosts the performance of the feature extraction models~\cite{islam2020learning, xu2021learning}. However, a study demonstrates that constant label smoothing lowers the utility of DNNs by degrading their refinement performance~\cite{singh2021dark}. Another study presents that ULS is incompatible with the semantic segmentation task and proposes spatially varying label smoothing (SVLS)~\cite{islam2021spatially} to confirm spatial variation among class regions. In this work, we adopt ULS and SVLS to control the learning utility as a curriculum scheme.

\subsection{Robustness}
The vast majority of research on robustness in for vision has focused on the critical issues of robustness to adversarial examples~\cite{szegedy2013intriguing,carlini2017adversarial,carlini2016defensive}, unknown unknowns~\cite{hendrycks2018deep}, and data poisoning~\cite{steinhardt2017certified}. Benchmark datasets for two other forms of robustness (corruption and perturbation) are developed to test the robustness of a classifier~\cite{hendrycks2019benchmarking}. The robustness enhancement is proved and validated on a diverse test set that contains the corrupted and perturbed images\cite{geirhos2018imagenet,zhang2019making}. In this work, we utilize the corruption and perturbation techniques to create a new test dataset to validate the robustness of our method.

\section{Background and preliminaries}

\begin{figure}[!hbpt]
\centering
\includegraphics[width=1\linewidth]{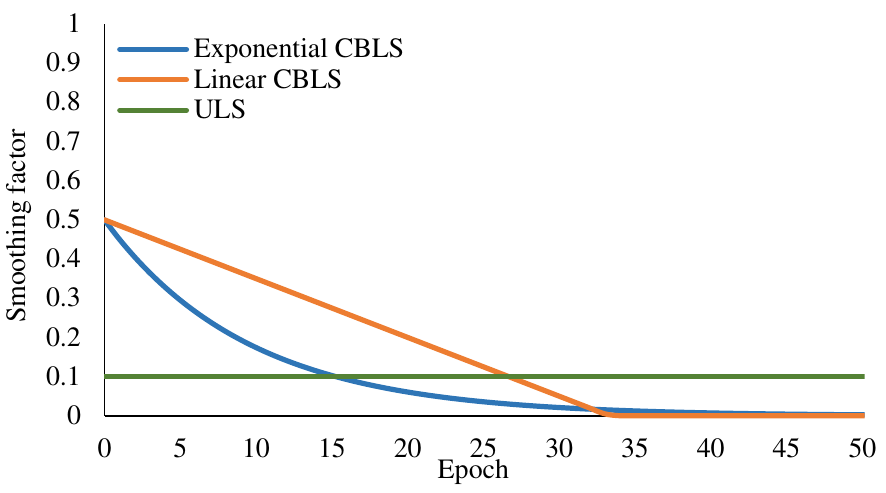}
\caption{\textbf{Illustration of ULS vs. Exponential CBLS vs. Linear CBLS.} The smoothing factor $\epsilon$ exponentially decreases in Exponential CBLS and linearly decreases in Linear CBLS. In ULS, the smoothing factor keeps constant.}
\label{fig:exponential_linear}
\end{figure}

\noindent\textbf{Uniform label smoothing (ULS)} Label smoothing (LS) is a regularization technique that improves the generalization and learning efficiency of DNNs by replacing one-hot labels with smoothed labels. It uniformly flattens the one-hot label ($T_{one-hot}$) by using a smoothing factor. Therefore we call it Uniform Label Smoothing (ULS) in this work. In $T_{one-hot}$, true class represents with ``1'' and the rest with ``0''. Then the smoothed label (soft label) $T_{ULS}$ is represented as
\begin{equation}
T_{ULS} = T_{one-hot}(1-\epsilon) + \epsilon/K
\label{equation:uls}
\end{equation}
where the number of classes $K$, the smoothing factor $\epsilon$ range of (0, 1) that decides smoothing strength and is always kept as a constant with training, as shown in Fig.~\ref{fig:exponential_linear}.

\noindent\textbf{Spatially varying label smoothing (SVLS)} LS smoothes the label uniformly, which is not compatible with semantic segmentation. Spatially Varying  Label Smoothing (SVLS)~\cite{islam2021spatially} is a soft labeling technique that captures the ambiguity and uncertainty about object boundaries in expert segmentation annotation. SVLS determines the probability of the target class based on neighboring pixels by designing an SVLS weight matrix, $w^{svls}$, with a Gaussian kernel  $k(x,y) = {\frac{1}{2\pi \sigma^2}}{e^{-\frac{\lvert \vec{x} \rvert^2}{2\sigma^2}}}$ with smoothing factor $\sigma$ set to 1. SVLS weight matrix is convolved across the one-hot encoding targets to obtain soft class probabilities, as shown in the equation below.

\begin{equation}
T_{SVLS} = w_{svls}(\sigma) \circledast  T_{one-hot}
\label{equation:svls}
\end{equation}

Similar to ULS, SVLS smoothing factor is also used to control smoothing strength and is kept constant during training. Changing the smoothing factor in a curriculum manner is still an unexplored area.

\noindent\textbf{Self-paced learning} Self-Paced Learning (SPL)~\cite{li2017self} incorporates a self-paced function $f(v)$ and a pace parameter $\gamma$ into the standard loss function. The total loss can be formulated as $\mathcal{L}_{total} = \sum_{i=1}^{n}(v_{i}{\mathcal{L}(t_{i}, y) + \gamma f(v)}$. The standard loss $\mathcal{L}(t_{i}, y)$ calculates the loss between the ground truth target $t_{i}$ and the predicted target $y$. The self-paced function is used to learn the weight variable $v$, which indicates whether the samples are easy or not. When pace parameter $\gamma$ is small, only ``easy'' samples with small losses are introduced into the training. As $\gamma$ increases, more ``difficulty'' samples with large loss are appended into training.

The weight variable $v$~\cite{li2017self} is dynamically updated during training. In our work, we implement confidence-aware paced learning based on the pre-decided samples bank.


\section{Paced-curriculum learning by label smoothing (P-CBLS)}
In this work, we design Paced-Curriculum by Label Smoothing (P-CBLS) using paced learning with ULS~\cite{szegedy2016rethinking} and SVLS~\cite{islam2021spatially} in a curriculum manner, as shown in Fig.~\ref{fig:main}. We build the confidence-aware sample bank sorted by sample difficulty and decrease the smoothing factors of ULS ($\epsilon$) and SVLS ($\sigma$) gradually to squeeze the label in a curriculum way where true probability value is modified from lower to higher during training epochs. It is worth noting that the proposed P-CBLS adds no additional trainable parameters, is generic, and can be used with any DNNs variant. Our novelty also lies in the task difficulty measurement strategy of sample-level difficulty and pixel-level to better cope with different task scenarios. Moreover, the investigation of the robustness of our P-CBLS helps to improve the generalization of the deep neural network. P-CBLS is described in more detail in the following sections.

\subsection{Curricula in ULS}
We design curricula in Uniform Label Smoothing (ULS) which aims to exponentially decrease the smoothing factor $\epsilon^c$ during training where higher values flatten the true probability heavily and reduce the learning utility. Therefore the model learns less information at the beginning epochs and gradually learns more information. The curriculum strategy is designed to anneal the smoothing factor $\epsilon^c$ with a decay rate of $\alpha$ ($\alpha < 1$) in every epoch during the training, as demonstrated in Fig.~\ref{fig:exponential_linear}. The attenuation of the smoothing factor can be implemented by using exponential or linear decrease. And we refer to them as exponential Curriculum by Label Smoothing (CBLS) and linear CBLS, respectively. Compared with linear CBLS, the smoothing factor of exponential CBLS decays more smoothly.

From Equation~\ref{equation:uls}, the curriculum smoothed target $T_{ULS}^{c}$ is formulated as

\begin{equation}
T_{ULS}^{c} = T_{one-hot}(1-\epsilon^c) + \epsilon^c/K 
\label{equation:CBLS_ULS}
\end{equation}
where, curriculum  smoothing factor $\epsilon^c = \alpha \epsilon^c$ and smoothing decay rate $\alpha$.

If $P$ is the predicted probability, then the CE loss with the curriculum smoothed target $T_{ULS}^{c}$  can be present as
\begin{equation}
\mathcal{L}_{CE}^{c} = - \sum_{i=1}^{K} T_{ULS(i)}^{c} log (P_{(i)})
\label{equation:CBLS_CE}
\end{equation}

\subsection{Curricula in SVLS}
We design a curriculum in Spatially Varying Label Smoothing (SVLS), decreasing boundary uncertainty in segmentation labels. For this purpose, we decrease the SVLS smoothing factor $\sigma$ from a higher to lower value over the training epoch to ensure less information to learn at the beginning of training and gradually increase the learning utility.

From Equation~\ref{equation:svls}, the curriculum smoothed target $T_{SVLS}^{c}$ is formulated as

\begin{equation}
T_{SVLS}^{c} = w_{svls}^{c}(\sigma^{c}) \circledast  T_{one-hot}
\label{equation:CBLS_SVLS}
\end{equation}
where, curriculum SVLS weight matrix $w_{svls}^{c}(\sigma^{c})$, smoothing factor $\sigma^{c} = \beta \sigma^{c}$ and smoothing decay rate $\beta$.

\subsection{Curricula in ULS and SVLS}
We design to implement the curricula in ULS and curricula in SVLS simultaneously. The curriculum smoothed target $T_{ULS+SVLS}^{c}$ is formulated as 

\begin{equation}
T_{ULS+SVLS}^{c} = w_{svls}^{c}(\sigma^{c}) \circledast  T_{ULS}^{c}(\epsilon^c)
\label{equation:CBLS_ULS_SVLS}
\end{equation}
where, $\sigma^{c}$ and $\epsilon^c$ are the curriculum smoothing factors for ULS and SVLS. The main difference between Equation~\ref{equation:CBLS_SVLS} and Equation~\ref{equation:CBLS_ULS_SVLS} is, $w_{svls}^{c}$ does convolution operation with $T_{ULS}^{c}$ rather than with $T_{one-hot}$. Fig.~\ref{fig:main} (a) shows an overview of labels and soft labels over the epochs for different training strategies where one-hot encoding label $T_{one-hot}$, ULS soft label $T_{ULS}^{c}$, SVLS soft label $T_{SVLS}^{c}$ and fused ULS \& SVLS soft label $T_{ULS+SVLS}^{c}$ in curricula manner.

\subsection{Confidence-aware paced learning}
To make sure easier information to train first, we design confidence-aware paced learning where harder samples are skipped at the beginning of the training and gradually introduced in later epochs. This is achieved by sorting the sample based on the confidence score from the baseline model to build a sample bank. A sample with high confidence score can be interpreted as the easier sample, and a sample with a low confidence score is a harder sample in the curriculum learning. The obtained easy samples and difficult samples are visualized in Fig.~\ref{fig:main} (b). To incorporate paced learning into curricula, we tune hyperparameters of the ratio of initial easy samples and ratio of epoch to introduce all the training samples. If the initial sample ratio $\lambda$, which is the ratio of initial sample size and the total number of samples when we start the training, the number of total epochs $E$, the epoch ratio $E_{all}$ which is the ratio of the epoch when introducing all training samples and the total epochs $E$, then the pace parameter $\mu$ (ratio of additional harder samples per epoch) can be formulated as
\begin{equation}
\mu =\frac{1.0-\lambda}{E_{all}\times E}
\end{equation}

In our experiments, we tune the initial sample ratio $\lambda$ and the epoch ratio $E_{all}$ to determine the pace parameter $\mu$. Our sample ratio $\lambda$ plus pace parameter $\mu$ have a similar meaning with the pace parameter $\gamma$ in~\cite{li2017self}.


To build our sample bank, we can sort the sample in two different techniques by utilizing the confidence score. 
\begin{itemize}
\item \textbf{Sample-wise} In sample-wise sorting, we can measure the confidence score for each sample and sort them from easy (higher confidence) to difficult (lower confidence). This can be applicable to multi-class classification and segmentation tasks. For multi-label classification, we take the average confidence of all instances in a sample and then sort all the samples based on averaged confidences, as shown in Fig.~\ref{fig:main} (b) (Sample-wise).

\item \textbf{Pixel-wise} Pixel-wise sample bank is only compatible with semantic segmentation task. The semantic segmentation task can be modeled as the dense pixel classification problem. Therefore, the sample bank can be constructed in pixels. In a pixel-wise sample-bank, we calculate the confidence of each pixel and sort them from easy to difficult pixels, as shown in Fig.~\ref{fig:main} (b) (Pixel-wise).
\end{itemize}

The details of calculating the sample-wise score for different tasks are
\begin{itemize}
\item \textbf{Workflow Classification} The sample-wise confidence score is obtained from the predicted probability of the true class.
\item \textbf{Tool Classification} The sample-wise confidence score is obtained by averaging the predicted probability of true classes. For the frames with no instrument, we average the predicted probability for all classes, which is a low value. 
\item \textbf{Tool Segmentation} The frame with no instrument has no learning value for the instrument segmentation task. Thus, the confidence score for such a frame is set to 0 directly. For the frame with instruments, we calculate the predicted probability of the true foreground classes and do the average for them. The background classes are not considered.
\item \textbf{Surgical Captioning} The sample-wise confidence score is obtained by averaging the predicted probability of true classes.
\end{itemize}

\subsection{P-CBLS}
We take our sample-wise P-CBLS (ULS) as an example to illustrate the algorithm of P-CBLS in Algorithm~\ref{algortihm_PCBLS}. Our proposed P-CBLS consists of CBLS, and confidence-aware paced learning. In CBLS, we decrease the smoothing factor $ \epsilon^c$ in a curriculum way every epoch. It can be formulated as $ \epsilon^c= \alpha \epsilon^c $, where the decay rate $\alpha$. Meanwhile, confidence-aware paced learning is incorporated. Specifically, the model parameter $w$ is updated every epoch. The model is trained with partial samples of size $L = \lambda \times N$ at the beginning, where the sample ratio $\lambda$. The sample size used to train the model increases by $(\mu \times e) \times N $ every epoch before the $(E_{all} \times E)$th epoch, where the total size of samples $N$, the pace parameter $\mu$, the total epochs $E$, and the epoch ratio $E_{all}$. When current epoch $e$ reaches the $(E_{all} \times E)$th epoch, the smoothing factor $ \epsilon^c$ is still being decayed, and the model is trained with all samples of size $N$until we complete the total epochs $E$.

\begin{algorithm}[!hbpt]
    \caption{\small{Algorithm of Sample-wise P-CBLS (ULS).}}
    \begin{algorithmic}[1]
        \label{algortihm_PCBLS}
        \small
        
        \STATE \textbf{Input:} The training dataset of varying size $L$, the total number of samples $N$, the sample ratio $\lambda$, 
        the pace parameter $\mu$. the total epochs $E$, the epoch ratio $E_{all}$, current epoch $e$, smoothing factor $\epsilon^c$, decay rate $\alpha$ ($\alpha < 1$)
        \STATE \textbf{Output:} Model parameter $w$\\
        \STATE \textbf{Training} Initialize $w^{*}$\\ 
        The initial training size $L = \lambda \times N$ \\
        $\mathbf{If}$\;$e$\textless $(E_{all} \times E)$\;$\mathbf{Then}$: \\  
        \hspace{0.25cm} $L = (\lambda + \mu \times e) \times N$ \\
        \hspace{0.25cm} $ \epsilon^c= \alpha \epsilon^c $\\
        \hspace{0.25cm} Update $w^{*}$ \\
        $\mathbf{Else}$\\
        \hspace{0.25cm} $L = N$ \\
        \hspace{0.25cm} $ \epsilon^c = \alpha \epsilon^c $\\
        \hspace{0.25cm} Update $w^{*}$ \\
        return $w = w^{*}$
    \end{algorithmic}
\end{algorithm}

We hope Algorithm~\ref{algortihm_PCBLS} helps to differentiate our approach from the CBS~\cite{sinha2020curriculumwH} approach. Our P-CBLS approaches depart from CBS~\cite{sinha2020curriculumwH} with several points: (1) The designed curricula are different where CBS~\cite{sinha2020curriculumwH} is designed based on feature smoothing and proposed CBLS is designed based on label smoothing. CBS~\cite{sinha2020curriculumwH} is focused on designing curricula based on smoothing the feature maps by adjusting the standard deviation $\delta$ of the Gaussian kernels. The information is gradually added, which leads to improvement in feature maps which Convolutional Neural Networks (CNNs) learn from. Our CBLS and P-CBLS smooth labels control the learning utility in a curriculum manner, which is an unexplored area. (2) Our P-CBLS approach further integrates sample-wise confidence-aware paced learning into CBLS (ULS) for the classification task and fuse pixel-wise confidence-aware paced learning and CBLS (SVLS) for the segmentation task. (3) The scope of experiments is different. CBS~\cite{sinha2020curriculumwH} is evaluated on the image classification task. Our approaches are evaluated for tasks ranging from image understanding tasks (multi-class workflow classification, multi-label tool classification, and segmentation) to the caption generation task. We also evaluate the robustness of our approaches.

\section{Experiments}
\subsection{Dataset}

\noindent\textbf{M2CAI16-Workflow Classification} is a public challenge dataset from MICCAI 2016 modeling and monitoring of computer-assisted interventions challenge~\cite{twinanda2016miccai}. The training dataset includes $27$ videos about cholecystectomy surgery. These videos are acquired at $25$fps and segmented into $8$ phases. $7$ videos from the training set are split as the validation set ($18723$ labeled frames), and the rest videos are split as the training set ($48854$ labeled frames). Each frame has a resolution of $1920 \times 1080$. We resize these frames into $250 \times 250$ following the work~\cite{jin2021temporal}.

\noindent\textbf{MICCAI17/18-Tool Classification} is a tool detection dataset built from the publicly available dataset MICCAI robotic instrument segmentation challenge 2017~\cite{allan20192017} and 2018~\cite{allan20202018}. We extract the instrument labels from segmentation annotation from these datasets. To maintain balance classes, the dataset is split into 1560/1244 images for train and validation. The images are resized to $224\times224$, and multiple instances can be present in an image with $8$ tool classes.

\noindent\textbf{Tool Segmentation} experiments are also conducted with MICCAI robotic instrument segmentation 2018~\cite{allan20202018}. A type-wise segmentation annotation is used in this work by following~\cite{gonzalez2020isinet}. The validation set includes the $2^{nd}$, $5^{th}$, $9^{th}$, and $15^{th}$ sequences ($596$ labeled frames). The training set includes the remaining sequences ($1639$ labeled frames). The frames are resized into half from the original resolution of $1024\times1280$. 

\noindent\textbf{Surgical Captioning} is also generated from the MICCAI robotic instrument segmentation challenge 2018~\cite{allan20202018}. The original training set includes 15 robotic nephrectomy operations obtained by the da Vinci X or Xi system. Each video sequence includes $149$ frames with a resolution of $1024 \times 1280$. After removing the $13^{th}$ sequence due to the fewer surgical activities, we split the training set ($14$ sequences) into two subsets following~\cite{xu2021class, xu2021learning}. The validation subset includes $1^{st}$, $5^{th}$, $16^{th}$ sequences ($447$ labeled frames). The training subset includes the remaining sequences ($1560$ labeled frames). The caption annotation is taken from~\cite{xu2021class}.

\subsection{Experiments and results}
In this section, we refer to a model trained with the standard way (i.i.d.) with cross-entropy (CE) loss and label smoothing as baseline and LS, respectively. Our method can be presented as CBLS (ULS), and CBLS (SVLS) based utilizing curriculum techniques using uniform LS and spatially varying LS. Both our variants can integrate with confidence-aware paced learning and refer to as P-CBLS (ULS) or P-CBLS (SVLS). We tune the hyper-parameters of the smoothing factor, decay, and initial pacing parameter and choose the best value for further experiments. We set smoothing factor $\epsilon$ of $0.1$ for all the LS experiments to maintain fair comparison. As the pacing function forms from the confidence score of the baseline, we calibrate the baseline using a well-known calibration technique, temperature scaling~\cite{guo2017calibration}.

\subsubsection{Classification}

We adopt two popular classification architectures ResNet50~\cite{he2016deep}, and DenseNet121~\cite{huang2017densely}, for multi-class and multi-label classification datasets of robot-assisted surgical workflow~\cite{twinanda2016miccai} and tool classification~\cite{allan20192017, allan20202018}. Multi-class workflow classification dataset trains on SGD optimizer, momentum of $0.9$, weight decay of $5e-3$, and learning rate decay of $0.1$ with an initial learning rate of $5e-3$ by following~\cite{jin2021temporal}. On the other hand, we follow previous work~\cite{wang2019graph} with the hyper-parameter setting where we use Adam optimizer with a learning rate $1e-4$ for the multi-label tool classification task. Other hyper-parameters for CBLS and P-CBLS are assigned as the initial smoothing factor of $0.5$, smoothing decay of $0.9$, initial sample ratio $\lambda$ of $0.6$, and epoch ratio for all samples $E_{all}$ of $0.4$.

The results are tabulated in Table \ref{table:classification}, where we report the accuracy for multi-class workflow classification and mean average precision (MAP) for multi-label tool classification. The proposed CBLS and P-CBLS have improved the performance by around $2\%$ in accuracy for DenseNet121 and MAP for ResNet50. Our CBLS approach shows the best results in the workflow classification task, and our P-CBLS approach achieves the best results in the tool classification task. Both CBLS and P-CBLS are our proposed methods. For either set of experiments, we do not tune any hyper-parameters. Therefore, we do not expect that P-CBLS will always get better results. The proper and suitable hyper-parameters of P-CBLS, including initial sample ratio $\lambda$ and epoch ratio for all samples $E_{all}$, may further boost the P-CBLS performance for different tasks or datasets.

\begin{table}[!hbpt]
\centering
\caption{\textbf{Workflow Classification, Tool Classification}. Classification accuracy and mean average precision (MAP) on M2CAI16-Workflow, MICCAI17/18-Tool dataset using baselines(ResNet50~\cite{he2016deep} and DenseNet121~\cite{huang2017densely}), ULS (smoothing factor of $0.1$), and our proposed CBLS (ULS) and P-CBLS (ULS). We use the same CBLS and P-CBLS hyper-parameters for a fair comparison.}
\scalebox{1}{
\begin{tabular}{cccc}
\hline
\multicolumn{2}{c|}{\multirow{2}{*}{Model}}                  & \multicolumn{1}{c|}{\textbf{Workflow}} & \textbf{Tool Class} \\
\multicolumn{2}{c|}{}                                        & \multicolumn{1}{c|}{Accuracy}          & MAP                 \\ \hline
\multicolumn{1}{c|}{\multirow{2}{*}{Baseline}} & Resnet50~\cite{he2016deep}    & 68.16                                  & 54.09               \\
\multicolumn{1}{c|}{}                          & LS~\cite{szegedy2016rethinking}          & 71.48                                  & 52.88               \\ \cline{1-1}
\multicolumn{1}{c|}{\multirow{2}{*}{Ours}}     & CBLS        & 71.66                                  & \textbf{54.92}      \\
\multicolumn{1}{c|}{}                          & P-CBLS       & \textbf{71.76}                         & 54.75               \\ \hline
\multicolumn{1}{c|}{\multirow{2}{*}{Baseline}} & Densenet121 & 67.75                                  & 53.58               \\
\multicolumn{1}{c|}{}                          & LS          & 68.65                                  & 55.42               \\ \cline{1-1}
\multicolumn{1}{c|}{\multirow{2}{*}{Ours}}     & CBLS        & 69.59                                  & \textbf{55.84}      \\
\multicolumn{1}{c|}{}                          & P-CBLS       & \textbf{70.43}                         & 55.75               \\ \hline
\end{tabular}
}
\label{table:classification}
\end{table}

\subsubsection{Segmentation}
For the tool segmentation task, we adopt two commonly used segmentation architectures of LinkNet34~\cite{chaurasia2017linknet, shvets2018automatic} and DeepLabv3+~\cite{chen2018encoder} and their implementation\footnote{https://github.com/ternaus/robot-surgery-segmentation}$^{,}$\footnote{https://github.com/MLearing/Pytorch-DeepLab-v3-plus}. The architectures are trained on Adam optimizer with a learning rate of $1e-4$ by following~\cite{shvets2018automatic}. The hyper-parameters for CBLS (ULS), CBLS (SVLS), and P-CBLS are the initial ULS smoothing factor ($\epsilon = 0.6,\ decay = 0.9$) and SVLS smoothing factor ($\sigma = 0.9,\ decay = 0.5$), initial pixel ratio ($\lambda = 0.8$) and epoch ratio for all samples ($E_{all} = 0.4$).

\begin{figure}[!hbpt]
\centering
\includegraphics[width=1\linewidth]{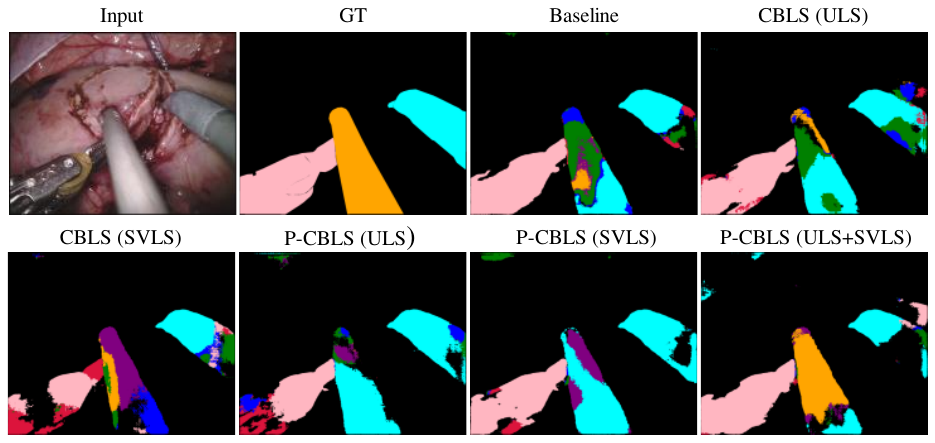}
\caption{\textbf{Visualization of the predicted mask of LinkNet34~\cite{chaurasia2017linknet} model based on different proposed approaches}. Bipolar forceps, prograsp forceps, monopolar curved scissors, and suction are indicated in pink, red, blue, and yellow. P-CBLS here specifically refers to pixel-wise paced learning.}
\label{fig:segmentation_result}
\end{figure}

The results are tabulated in Table \ref{table:segmentation}, where we report the mean IoU and mean Dice. The proposed various CBLS and P-CBLS have improved the performance at least by $3\%$ in Dice and $1\%$ in IoU for LinkNet34. For DeepLabv3+, the proposed methods obtain around $1\%$ improvement in Dice. In addition to significantly improving the IoU and Dice of the baseline CE approach, we see that training both model architectures using our CBLS (SVLS) outperforms our CBLS (ULS) by a good margin. The improvement in these two metrics suggests that our CBLS (SVLS) makes models better at capturing the ambiguity about the object contours. By fusing the CBLS (ULS) and CBLS (SVLs), we can see further performance improvements. Based on these findings, we conduct the same experimental design by incorporating pixel-wise confidence-aware paced learning. For LinkNet34, P-CBLS (ULS+SVLS) obtains the best performance. In comparison, P-CBLS (ULS) attains the best performance for DeepLabv3+. We attribute it to the fact that DeepLabv3+ converges faster than LinkNet34 usually. Therefore, the selected hyper-parameters of P-CBLS in our fixed setting cannot further contribute to improving the performance of the DeepLabv3+. The predicted masks generated by our proposed models are visualized in Fig.~\ref{fig:segmentation_result}. We can observe that for the ``suction" instrument, which is indicated by the yellow color in Ground Truth (GT) image, the traditional CE approach fails to predict it. For the ``monopolar curved scissors" instrument, which is represented by blue color in the GT image, the prediction of the CE approach is still largely wrong. Our various CBLS versions, including CBLS (ULS), CBLS (SVLS), P-CBLS (ULS), and P-CBLS (SVLS), show better performance on the ``monopolar curved scissors" instrument. Our P-CBLS (ULS+SVLS) shows superior performance on the ``suction" instrument.

\begin{figure*}[!hbpt]
\centering
\includegraphics[width=1\linewidth]{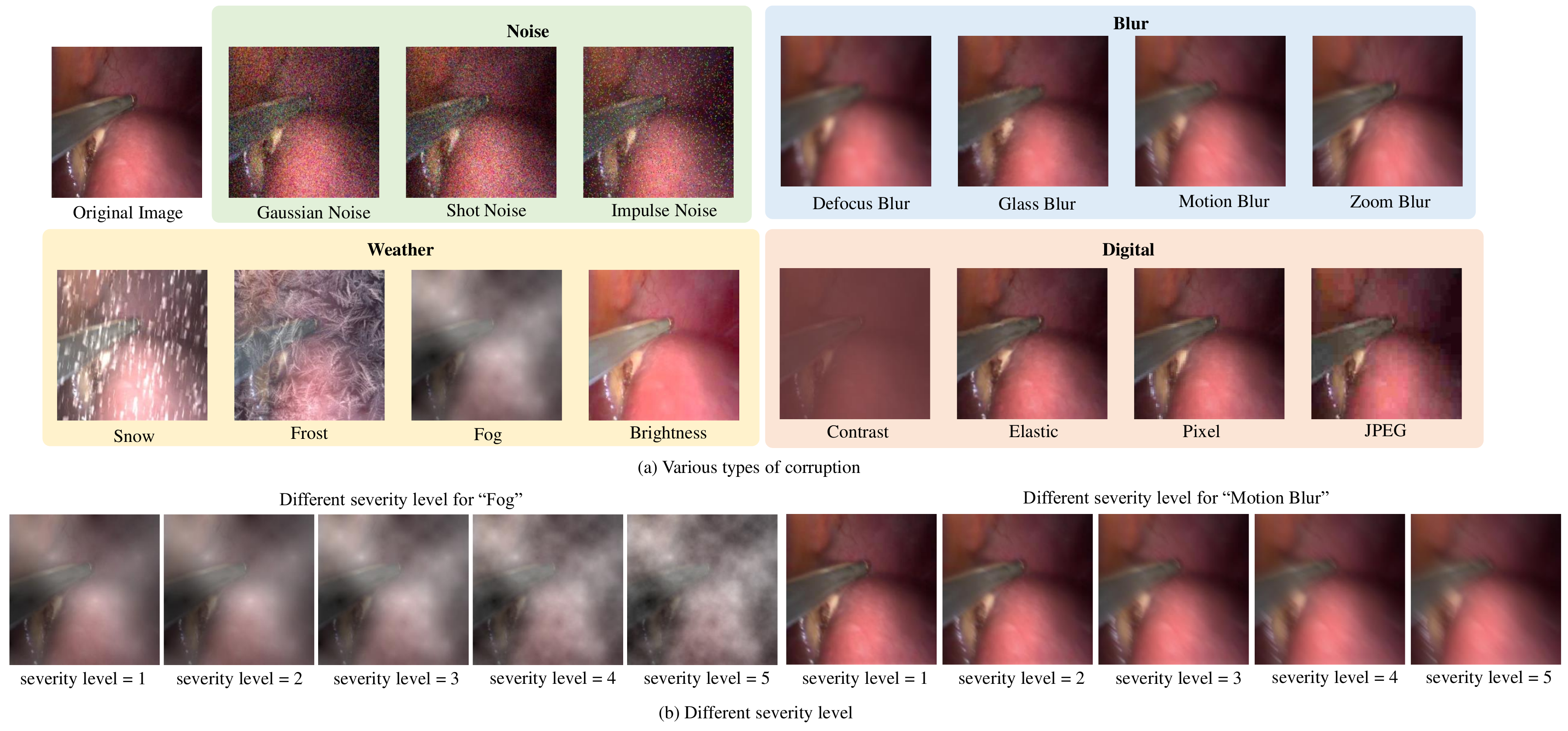}
\caption{\textbf{Visualization of various types of corruption and different severity levels on M2CAI16 Workflow dataset}. (a) Four types of corruption with severity level = 3, including noise, blur, weather, and digital, which are from the following filters ``Gaussian, Shot, Impulse", ``Defocus, Glass, Motion, Zoom", ``Snow, Frost, Fog, Bright", and ``Contrast, Elastic, Pixel, JPEG" respectively. (b) Five different severity levels.}
\label{fig:corrupted_images}
\end{figure*}

\begin{table}[!hbpt]
\centering
\caption{\textbf{Tool Segmentation.} mean IoU and mean Dice are reported on the tool Segmentation dataset for our CBLS variants over baselines of label smoothing (LS) and cross-entropy (CE) loss using LinkNet34~\cite{chaurasia2017linknet} and fix~\cite{chen2018encoder}.}
\scalebox{1.0}{
\begin{tabular}{cccccc}
\hline
\multicolumn{2}{c|}{\multirow{2}{*}{Model}}                        & \multicolumn{2}{c|}{LinkNet34}             & \multicolumn{2}{c}{DeepLabv3+}     \\
\multicolumn{2}{c|}{}                                              & IoU            & \multicolumn{1}{c|}{Dice} & IoU            & Dice           \\ \hline
\multicolumn{1}{c|}{\multirow{4}{*}{Baseline}} & CE                & 44.04          & 53.56                     & 44.62          & 54.67          \\
\multicolumn{1}{c|}{}                          & LS~\cite{szegedy2016rethinking}                & 42.75          & 50.78                     & 44.15          & 53.44          \\
\multicolumn{1}{c|}{}                          & Online LS~\cite{zhang2020delving}         & 43.70          & 51.94                     & 45.93          & 56.38          \\
\multicolumn{1}{c|}{}                          & SCE~\cite{wang2019symmetric}               & 47.05          & 56.10                     & 45.93          & 54.97          \\ \hline
\multicolumn{1}{c|}{\multirow{6}{*}{Ours}}     & CBLS (ULS)        & 47.35          & 54.95                     & 45.24          & 54.53          \\
\multicolumn{1}{c|}{}                          & CBLS (SVLS)       & 48.91          & 59.92                     & 47.93          & 57.31          \\
\multicolumn{1}{c|}{}                          & CBLS (ULS+SVLS)   & 50.63          & 61.14                     & 47.92          & 55.21          \\
\multicolumn{1}{c|}{}                          & P-CBLS (ULS)      & 48.71          & 57.54                     & \textbf{48.33} & \textbf{58.08} \\
\multicolumn{1}{c|}{}                          & P-CBLS (SVLS)     & 53.23          & 64.94                     & 45.28          & 55.23          \\
\multicolumn{1}{c|}{}                          & P-CBLS (ULS+SVLS) & \textbf{54.71} & \textbf{65.65}            & 46.18          & 55.50          \\ \hline
\end{tabular}}
\label{table:segmentation}
\end{table}

\subsubsection{Captioning}
For the image to caption generation, we use Mesh-Transformer (M2T)\footnote{https://github.com/aimagelab/meshed-memory-transformer}~\cite{cornia2020meshed} and adopt the same parameters and object features from~\cite{xu2021class}. All models are evaluated using four metrics for image captioning, namely BLEU-n~\cite{papineni2002bleu}, ROUGE~\cite{lin2004rouge}, METEOR~\cite{banerjee2005meteor}, CIDEr~\cite{vedantam2015cider}. We apply our curriculum methods on the M2T~\cite{cornia2020meshed}
and compared with the original M2T~\cite{cornia2020meshed}, X-LAN~\cite{pan2020x}, and image captioning models from~\cite{xu2021learning}.
We use initial smoothing factor $\epsilon$ of $0.1$, decay rate $\alpha$ of $0.95$, and initial sample ratio $\lambda$ of $0.9$.

\begin{figure}[!hbpt]
\centering
\includegraphics[width=1\linewidth]{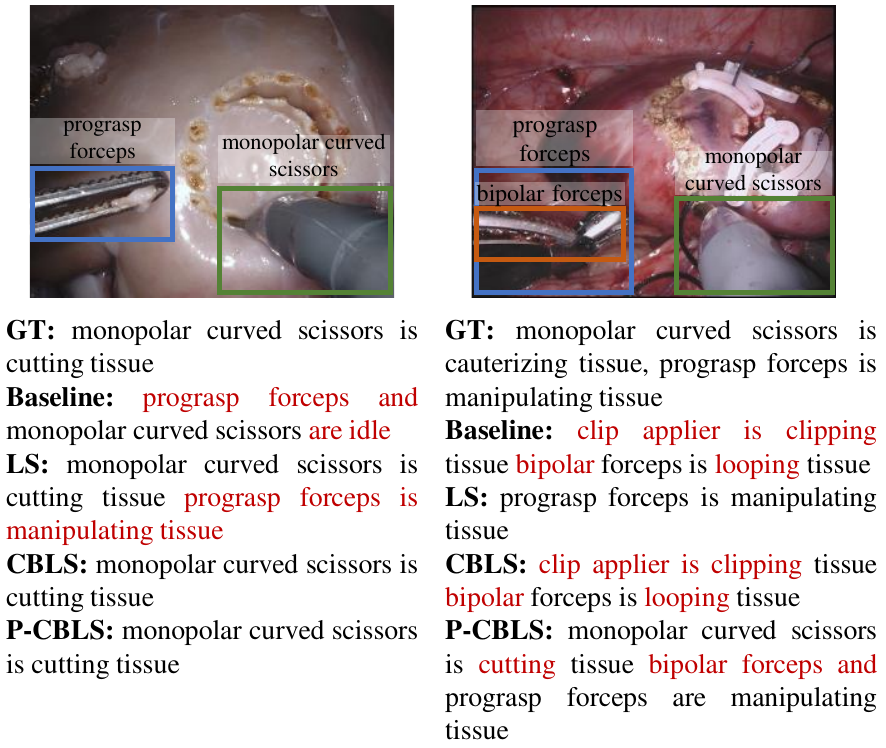}
\caption{\textbf{Visualization of the predicted caption with our proposed CBLS, P-CBLS, LS, and baseline.} The red text indicates the difference part between the predicted caption and ground truth.}
\label{fig:caption_result}
\end{figure}

The caption prediction of our curriculum-based networks is visualized in Figure \ref{fig:caption_result}. Our methods show superiority over other state-of-the-art methods on the M2CAI-2018 captioning dataset. As shown in TABLE~\ref{result_captioning}, the proposed P-CBLS has improved the performance by around $0.01$ in BLEU-1 and METEOR, around $0.02$ in ROUGE, and around $0.7$ in CIDEr when compared with~\cite{xu2021learning}. Our P-CBLS presents better performance on BLEU-4, ROUGE, and CIDEr.

\begin{table}[!hbpt]
\centering
\caption{\textbf{Surgical Captioning}. BLEU-n~\cite{papineni2002bleu}, METEOR~\cite{banerjee2005meteor}, ROUGE~\cite{lin2004rouge} and CIDEr~\cite{vedantam2015cider} on the surgical captioning dataset for network architecture trained normally, trained using ULS with smoothing factor of $0.1$, and trained using our proposed CBLS and P-CBLS. Compared with baseline architectures: X-LAN~\cite{pan2020x}, Xu et al.~\cite{xu2021class}, M2T~\cite{cornia2020meshed}, we show significant improvements using our proposed methods based on the M2T~\cite{cornia2020meshed}.}
\label{result_captioning}
\scalebox{0.9}{
\begin{tabular}{cc|c|c|c|c}
\hline
\multicolumn{2}{c|}{\textbf{Model}}                       & \textbf{BLEU-4} & \textbf{METEOR} & \textbf{ROUGE}  & \textbf{CIDEr}  \\ \hline
\multicolumn{1}{c|}{\multirow{4}{*}{Baseline}} & X-LAN~\cite{pan2020x}    & 38.85          & 34.84          & 56.42          & 205.99          \\ 
\multicolumn{1}{c|}{}                          & Xu et al~\cite{xu2021class} & 44.52          & 35.32          & 60.17          & 265.24          \\ 
\multicolumn{1}{c|}{}                          & M2T~\cite{cornia2020meshed}      & 42.29          & 35.67          & 58.95          & 274.78          \\ 
\multicolumn{1}{c|}{}                          & LS (M2T) & 46.21          & 35.45          & 60.53          & 296.04          \\ \hline
\multicolumn{1}{c|}{\multirow{2}{*}{Ours}}     & CBLS     & 45.07          & \textbf{36.52} & 60.25          & 309.65          \\ 
\multicolumn{1}{c|}{}                          & P-CBLS    & \textbf{47.46} & 35.67          & \textbf{62.63} & \textbf{340.67} \\ \hline
\end{tabular}}
\end{table}


\subsection{Robustness}
\begin{figure}[!hbpt]
\centering
\includegraphics[width=1\linewidth]{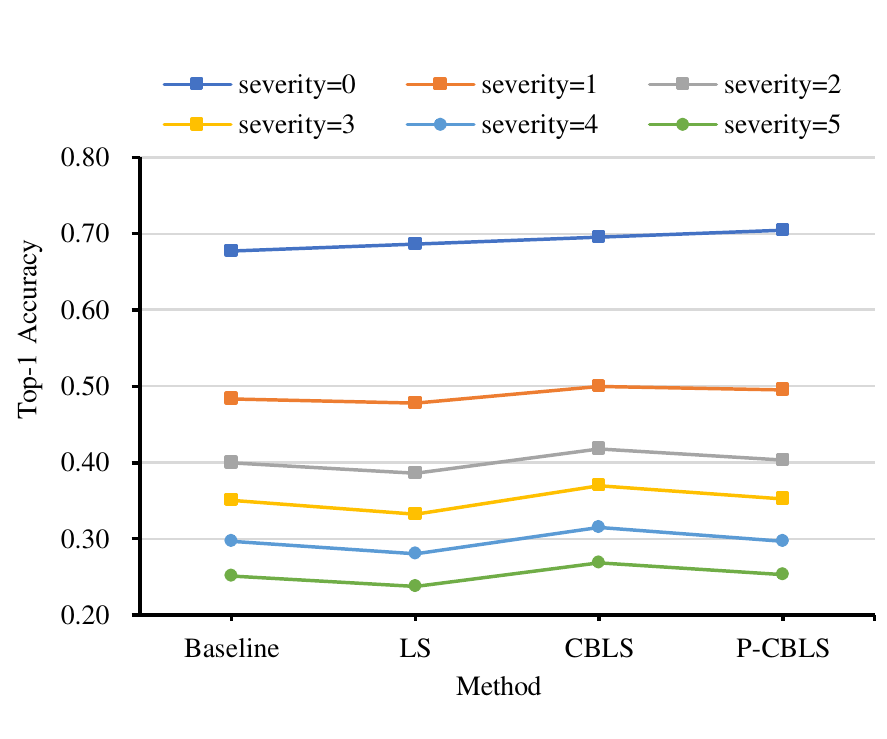}
\caption{\textbf{Robustness performance}. Performance of DenseNet121 for our CBLS and P-CBLS compared to baselines across various severity levels of corruption on the Workflow classification dataset.}
\label{fig:robustness}
\end{figure}

A more quantitative approach for analyzing the robustness of a model is its response to manually generated corruption and perturbation to the images\cite{zhang2019making}. Therefore, we design four types of corruption, including noise, blur, weather, and digital, which are from the following filters ``Gaussian, Shot, Impulse", ``Defocus, Glass, Motion, Zoom", ``Snow, Frost, Fog, Bright", and ``Contrast, Elastic, Pixel, JPEG" respectively to validate the robustness performance of the baselines, and our proposed CBLS and P-CBLS, as shown in Fig.~\ref{fig:corrupted_images}. When the model is more robust to corruption and perturbation, it will show higher accuracy. As the severity increases, if the model can preserve performance, then it can be regarded as more robust\cite{hendrycks2019benchmarking}. Results are demonstrated in TABLE~\ref{table:robustness_experiments}. Our approaches obtain competitive results overall. Although CBLS yields 0.5, 0.6, and 0.2 lower than the baseline for Blurred-Glass, Blurred-Motion, and Weather-Fog, our CBLS and P-CBLS significantly outperform most corruption types. In our experiments, each corruption type integrates $5$ different corruption severity levels. For Blurred-Glass, Blurred-Motion, Weather-Fog, overhigh corruption severity levels may make the image deviate too much from the original image, which makes the corrupted image illegible. Our approach aims to control the model learning utility from lower to higher in the curriculum manner. Compared to the baseline, our approaches may be more sensitive to such corruption types, which causes too much deviation from the original image because the model with only a lower learning utility in the early stages of training may face frustration when encountering such indistinguishable and difficult images. We also plot the performance degradation over the five severity levels of the corruption in Fig.~\ref{fig:robustness}. The results suggest that LS and baseline are remarkably degraded with the increase in severity, whereas CBLS preserves better performance.

\begin{table*}[!hbpt]
\centering
\caption{\textbf{Performance of our CBLS and P-CBLS over baselines under various types of corruption on M2CAI16 Workflow dataset with DenseNet121.} The results are obtained from five different severity levels and average them for each corruption.}
\scalebox{.85}{
\begin{tabular}{c|ccc|cccc|cccc|cccc|c}
\hline
         & \multicolumn{3}{c|}{Noise}                                                                 & \multicolumn{4}{c|}{Blur}                                                                       & \multicolumn{4}{c|}{Weather}                                                                  & \multicolumn{4}{c|}{Digital}                                                                      &                        \\ \cline{1-16}
Model    & \multicolumn{1}{c|}{Gauss.} & \multicolumn{1}{c|}{Shot}  & Impulse & \multicolumn{1}{c|}{Defocus} & \multicolumn{1}{c|}{Glass} & \multicolumn{1}{c|}{Motion} & Zoom  & \multicolumn{1}{c|}{Snow}  & \multicolumn{1}{c|}{Frost} & \multicolumn{1}{c|}{Fog}   & Bright & \multicolumn{1}{c|}{Contrast} & \multicolumn{1}{c|}{Elastic} & \multicolumn{1}{c|}{Pixel} & JPEG  & \multirow{-2}{*}{Mean} \\ \hline
Baseline & \multicolumn{1}{c|}{32.5}                          & \multicolumn{1}{c|}{33.6} & 29.3   & \multicolumn{1}{c|}{37.4}   & \multicolumn{1}{c|}{\textbf{41.9}} & \multicolumn{1}{c|}{\textbf{42.6}}  & 42.8 & \multicolumn{1}{c|}{29.4} & \multicolumn{1}{c|}{32.5} & \multicolumn{1}{c|}{\textbf{43.0}} & 53.5  & \multicolumn{1}{c|}{33.8}    & \multicolumn{1}{c|}{55.7}   & \multicolumn{1}{c|}{58.2} & 48.8 & 41.0                  \\ 
LS       & \multicolumn{1}{c|}{30.7}                          & \multicolumn{1}{c|}{31.4} & 30.8   & \multicolumn{1}{c|}{34.2}   & \multicolumn{1}{c|}{38.8} & \multicolumn{1}{c|}{38.7}  & 43.6 & \multicolumn{1}{c|}{26.4} & \multicolumn{1}{c|}{29.8} & \multicolumn{1}{c|}{42.2} & 54.7  & \multicolumn{1}{c|}{36.9}    & \multicolumn{1}{c|}{55.5}   & \multicolumn{1}{c|}{58.4} & 48.2 & 40.0                  \\
CBLS     & \multicolumn{1}{c|}{33.8}                          & \multicolumn{1}{c|}{34.8} & 31.4   & \multicolumn{1}{c|}{37.7}   & \multicolumn{1}{c|}{41.4} & \multicolumn{1}{c|}{42.0}  & \textbf{47.8} & \multicolumn{1}{c|}{\textbf{32.4}} & \multicolumn{1}{c|}{\textbf{33.3}} & \multicolumn{1}{c|}{42.8} & \textbf{57.6}  & \multicolumn{1}{c|}{\textbf{38.4}}    & \multicolumn{1}{c|}{\textbf{57.5}}   & \multicolumn{1}{c|}{\textbf{60.8}} & 50.0 & \textbf{42.8}                 \\ 
P-CBLS   & \multicolumn{1}{c|}{\textbf{35.0}}                          & \multicolumn{1}{c|}{\textbf{35.7}} & \textbf{32.3}   & \multicolumn{1}{c|}{\textbf{38.3}}   & \multicolumn{1}{c|}{40.0} & \multicolumn{1}{c|}{41.4}  & 44.8 & \multicolumn{1}{c|}{28.6} & \multicolumn{1}{c|}{27.7} & \multicolumn{1}{c|}{42.0} & 55.2  & \multicolumn{1}{c|}{37.5}    & \multicolumn{1}{c|}{\textbf{57.5}}   & \multicolumn{1}{c|}{57.9} & \textbf{52.6} & 4.18                  \\ \hline
\end{tabular}}
\label{table:robustness_experiments}
\end{table*}


\begin{figure*}[!hbpt]
\centering
\includegraphics[width=1.0\linewidth]{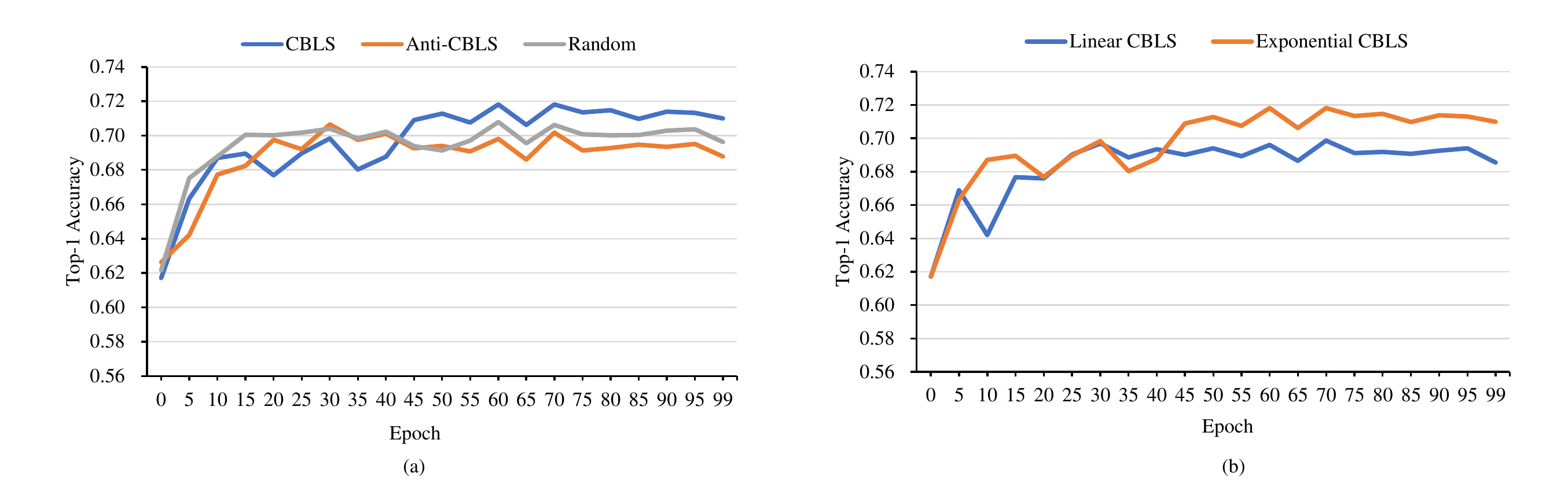}
\caption{\textbf{Different training paradigms to change the smoothing factor with ResNet50 on the M2CAI16-Workflow dataset.} (a) CBLS vs. Anti CBLS vs. Random. (b) Exponential CBLS vs. Linear CBLS.}
\label{fig:cbls_anticbls_random}
\end{figure*}

\begin{figure*}[!hbpt]
\centering
\includegraphics[width=1.0\linewidth]{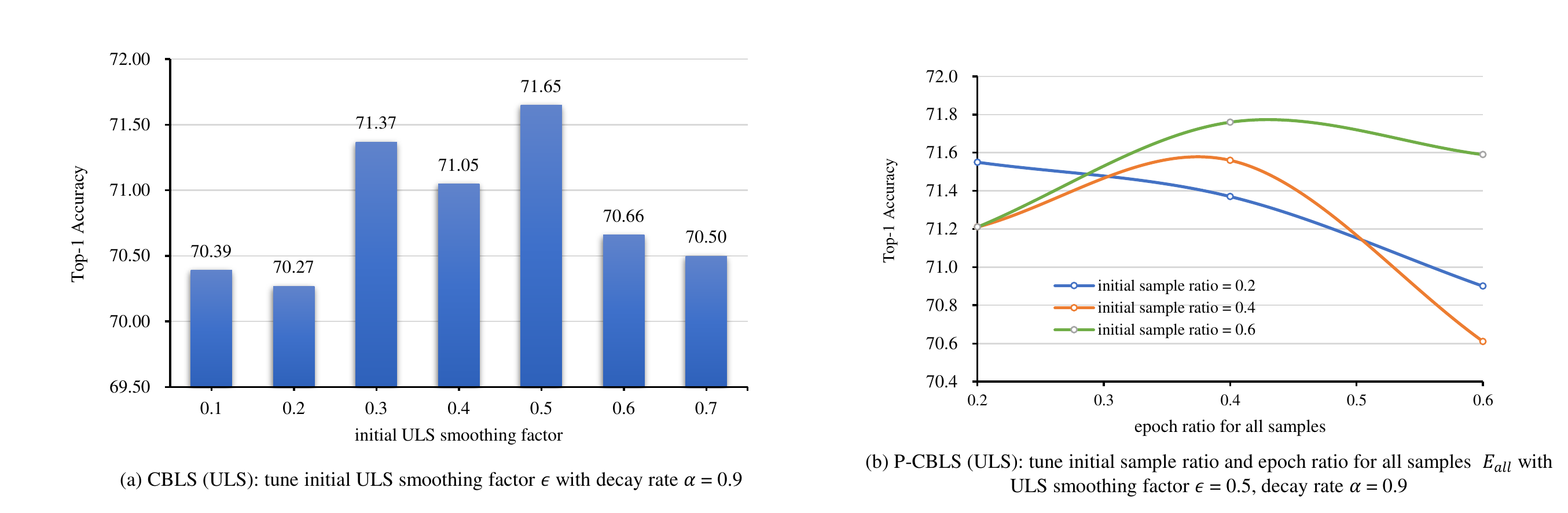}
\caption{\textbf{Ablation study of the hyperparameters of our proposed approaches on the M2CAI16 Workflow Classification dataset with ResNet50 model.} (a) hyperparameters of CBLS. (b) hyperparameters of P-CBLS.}
\label{fig:ablation_MAP_Accuracy}
\end{figure*}

\subsection{Ablation study}

\subsubsection{\textbf{Curriculum vs. anti-curriculum vs. random}}
We design three different training paradigms to change the smoothing factor: \textbf{CBLS} (initial ULS smoothing factor of 0.5, the decay rate of 0.9. The minimum of ULS smoothing factor is restricted to 0), \textbf{Anti-CBLS} (initial ULS smoothing factor of 0.005, the decay rate of 1.1. The maximum of ULS smoothing factor is restricted to 0.5) and \textbf{Random} (ULS smoothing factor is from the random value within the range of (0, 0.5) during training) on M2CAI16-Workflow dataset, to demonstrate the superiority of our CBLS approach. The results are shown in Fig.~\ref{fig:cbls_anticbls_random} (a). It is interesting to see that Anti-CBLS performs worse than CBLS and the random way to choose the smoothing factor.

\subsubsection{\textbf{Exponential CBLS vs. linear CBLS}}
We study the attenuation of smoothing in an exponential way (initial ULS smoothing factor of 0.5, decay rate of 0.9) and a linear way (initial ULS smoothing factor of 0.5, decay rate of 0.015) on the M2CAI16-Workflow dataset. And we refer to them as Exponential CBLS and Linear CBLS, respectively. We observed that the Exponential CBLS outperforms Linear CBLS by a large margin, as demonstrated in Fig.~\ref{fig:cbls_anticbls_random} (b). We also show the comparison between exponential CBLS and linear CBLS based on the Tool Segmentation dataset, as demonstrated in TABLE~\ref{table:ablation_segmentation}. Exponential CBLS has better performance than linear CBLS.

\subsubsection{\textbf{Sample ranking and calibration}}
We analyze the different sample ranking approaches and report the corresponding calibration performance on the M2CAI16-Workflow dataset. Expected Calibration Error (ECE)~\cite{naeini2015obtaining} is a common metric often used to measure calibration, and it is computed as the difference between the accuracy and predicted confidences calculated over several bins. We employ two calibration approaches in our work: temperature scaling (TS)~\cite{guo2017calibration} and label smoothing (LS)~\cite{muller2019does}. In P-CBLS, we calculate the confidence score to evaluate the difficulty of the sample based on the baseline model, which is trained using cross-entropy loss. In ls$\_$P-CBLS, we use the model trained with the LS technique (ULS smoothing factor $\epsilon$ = $0.1$). In ts$\_$P-CBLS, we use the standard model but with the TS technique (T = $2.63$ for ResNet50, and T = $2.53$ for DenseNet121). The results in Table \ref{table:ts_PCBLS_and_ls_PCBLS} show that our P-CBLS, ts$\_$P-CBLS, and ls$\_$P-CBLS approaches attain lower calibration errors than the baseline.

\begin{table}[!hbpt]
\centering
\caption{\textbf{Sample-wise difficulty ranking study.} To obtain the samples bank sorted by ``easiness" for the paced learning, we utilize three different kinds of models: (1) the standard (i.i.d.) training; (2) calibrated model with temperature scale technique (T=2.63 for ResNet50, T=2.53 for DenseNet121); (3) the calibrated model with constant label smoothing. Based on the three different sample banks, the model trained with P-CBLS is referred to as P-CBLS, ts$\_$P-CBLS, ls$\_$P-CBLS, respectively. We report the classification performance and calibration performance with the ECE for two network architectures: ResNet50 and DenseNet121.}
\scalebox{0.9}{
\begin{tabular}{cc|c|l}
\hline
\multicolumn{2}{c|}{\textbf{Model}}                      & \textbf{Accuracy} & \multicolumn{1}{c}{\textbf{ECE}$\downarrow$} \\ \hline
\multicolumn{1}{c|}{Baseline}              & ResNet50    & 68.16             & 0.2063                           \\ \hline
\multicolumn{1}{c|}{\multirow{3}{*}{Ours}} & P-CBLS       & 71.76             & 0.0955                           \\  
\multicolumn{1}{c|}{}                      & ts$\_$P-CBLS    & \textbf{71.80}    & 0.1630                           \\ 
\multicolumn{1}{c|}{}                      & ls$\_$P-CBLS    & 70.67             & \textbf{0.0726}                  \\ \hline
\multicolumn{1}{c|}{Baseline}              & DenseNet121 & 67.75             & 0.2070                           \\ \hline
\multicolumn{1}{c|}{\multirow{3}{*}{Ours}} & P-CBLS       & 70.43             & \textbf{0.1115}                  \\ 
\multicolumn{1}{c|}{}                      & ts$\_$P-CBLS    & 69.32             & 0.1957                           \\ 
\multicolumn{1}{c|}{}                      & ls$\_$P-CBLS    & \textbf{70.66}    & 0.1735                           \\ \hline
\end{tabular}}
\label{table:ts_PCBLS_and_ls_PCBLS}
\end{table}


\subsubsection{\textbf{Sample-wise P-CBLS vs. pixel-wise P-CBLS}}
The comparison between sample-wise P-CBLS and pixel-wise P-CBLS is shown in TABLE~\ref{table:ablation_segmentation}.


\begin{table}[!hbpt]
\centering
\caption{Segmentation results from different training strategies based on LinkNet34~\cite{chaurasia2017linknet} model.}
\scalebox{0.9}{
\begin{tabular}{c|c|c}
\hline
                 & IoU    & Dice   \\ \hline
Linear CBLS      & 44.49 & 53.52 \\ 
Exponential CBLS & \textbf{47.35} & \textbf{54.95} \\ 
sample wise P-CBLS & 46.65 &  \textbf{57.82}
 \\ 
pixel wise P-CBLS     & \textbf{48.71} & 57.54 \\ \hline
\end{tabular}}
\label{table:ablation_segmentation}
\end{table}

\subsubsection{\textbf{Hyper-parameters of P-CBLS (ULS)}}
We investigate the effect of different initial ULS smoothing factor $\epsilon$, decay rate $\alpha$, initial sample ratio $\lambda$, and epoch ratio for all samples $E_{all}$ for ResNet50 on M2CAI16-Workflow Classification dataset (see Fig.~\ref{fig:ablation_MAP_Accuracy}). The initial ULS smoothing factor $\epsilon$ of $0.5$ generate the best accuracy, as shown in Fig.~\ref{fig:ablation_MAP_Accuracy} (a). Therefore, we fine-tune the decay rate $\alpha$ with range $\alpha \in [0.8, 0.95]$ at a step of $0.05$ with initial ULS smoothing factor $\epsilon$ of $0.5$. We did not see any further improvement from the fine-tuning operation. Next, we use the initial ULS smoothing factor $\epsilon$ of $0.5$ and decay rate $\alpha$ of 0.9 for P-CBLS (ULS) experiments where we tune the initial sample ratio $\lambda$ with range $\lambda \in [0.2, 0.6]$ at a step of $0.2$ and the epoch ratio for all samples $E_{all}$ with range $E_{all} \in [0.2, 0.6]$ at a step of $0.2$. The initial sample ratio $\lambda$ of $0.6$ and epoch ratio for all samples $E_{all}$ of $0.4$ generate the best accuracy, as shown in Fig.~\ref{fig:ablation_MAP_Accuracy} (b). 

\subsubsection{\textbf{Validation on computer vision datasets}}

In TABLE~\ref{table:cv}, we report the accuracy for three common computer vision datasets, including CIFAR10, CIFAR100, and Tiny-ImageNet. Compared to CE, LS~\cite{szegedy2016rethinking}, Online LS~\cite{zhang2020delving} and Disturb Label~\cite{xie2016disturblabel}, our CBLS shows superior results.

\begin{table}[!t]
\centering
\caption{Classification results of our CBLS and other baselines (CE, LS, Online LS~\cite{zhang2020delving} and Disturb Label~\cite{xie2016disturblabel}) on computer vision datasets.}      
\scalebox{0.9}{
\begin{tabular}{c|c|c|c|c}
\hline
                             &               & \multirow{2}{*}{\begin{tabular}[c]{@{}c@{}}\textbf{CIFAR}\\      \textbf{10}\end{tabular}} & \multirow{2}{*}{\begin{tabular}[c]{@{}c@{}}\textbf{CIFAR}\\      \textbf{100}\end{tabular}} & \multirow{2}{*}{\begin{tabular}[c]{@{}c@{}}\textbf{Tiny}\\      \textbf{ImageNet}\end{tabular}} \\ \cline{1-2}
Model                        & Method        &                                                                          &                                                                           &                                                                               \\ \hline
\multirow{5}{*}{\rotatebox{90}{ResNet50}}    & CE            & 84.14                                                                    & 66.05                                                                     & 58.94                                                                         \\ 
                             & LS~\cite{szegedy2016rethinking}            & 84.70                                                                    & 66.69                                                                     & 59.90                                                                         \\ 
                             & Online LS~\cite{zhang2020delving}     & 82.08                                                                    & 66.20                                                                     & 59.89                                                                         \\ 
                             & Disturb Label~\cite{xie2016disturblabel} & 81.22                                                                    & 66.17                                                                     & 58.84                                                                         \\ 
                             & CBLS (Ours)          & \textbf{88.59}                                                           & \textbf{67.26}                                                            & \textbf{60.68}                                                                \\ \hline
\multirow{5}{*}{\rotatebox{90}{DenseNet121}} & CE            & 78.11                                                                    & 68.09                                                                     & 63.53                                                                         \\  
                             & LS~\cite{szegedy2016rethinking}            & 80.67                                                                    & 69.19                                                                     & 64.02                                                                         \\ 
                             & Online LS~\cite{zhang2020delving}     & 75.83                                                                    & 68.60                                                                     & 63.95                                                                         \\  
                             & Disturb Label~\cite{xie2016disturblabel} & 71.81                                                                    & 58.31                                                                     & 62.99                                                                         \\ 
                             & CBLS (Ours)          & \textbf{84.04}                                                           & \textbf{69.50}                                                            & \textbf{64.66}                                                                \\ \hline
\end{tabular}}
\label{table:cv}
\end{table}


\section{Discussion and conclusion}
Previous research has attempted to improve surgical recognition tasks by incorporating additional modules or making the technique computationally intensive. However, this has slowed prediction and limited its use in real-time robotic applications. The novel Paced-Curriculum By Label Smoothing (P-CBLS) method is proposed to improve the performance and generalization of deep neural networks (DNNs) without adding additional training parameters, which learns the samples from easy to complex with the gradually annealed smoothing factor. Extensive experiments on multiple robotic vision datasets for surgical recognition tasks demonstrate the effectiveness and robustness of our proposed CBLS and P-CBLS on different models. Our results and analyses suggest that curriculum learning can be developed by smoothing labels and controlling learning utility over the epochs. We also find that CBLS is better than P-CBLS for highly corrupted images with higher severity. As CBLS controls true probability during the training, it does not introduce additional parameters in the model. Therefore, CBLS and P-CBLS are model and task-agnostic curriculum learning strategies and are simple yet effective for many applications. Tuning the initial smoothing factor and decay rate in our CBLS approach may produce better predictions than our reported results using common hyper-parameters for both architectures. We investigate that after tuning, we can obtain $56.04$ in CBLS and $55.94$ in P-CBLS, which are higher than the reported results in TABLE~\ref{table:classification} for the tool classification task. In future work, the confidence information can be integrated with label smoothing to re-weight the smoothing factor by class instance during CBLS training. It is also interesting to investigate the amalgamation of class distribution knowledge with the smoothing factor to design a curriculum scheme for the long-tailed dataset.


\bibliographystyle{IEEEtran}
\bibliography{sample}

\begin{IEEEbiography}[{\includegraphics[width=1in,height=1.25in,clip,keepaspectratio]{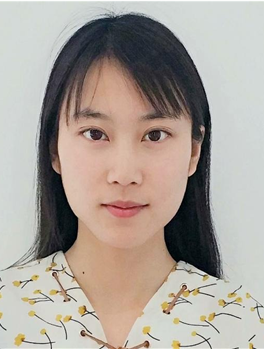}}]{Mengya Xu}
received a B.Eng. degree in information engineering from Soochow University, Suzhou, China, in 2018 and an M.Sc. degree in electrical and computer engineering from the National University of Singapore, Singapore. She is currently pursuing a Ph.D. degree with the Department of Biomedical Engineering, National University of Singapore, Singapore. Her research focuses on vision-language multimodality-based surgical scene understanding, supervised by Prof. Hongliang Ren.
\end{IEEEbiography}

\begin{IEEEbiography}[{\includegraphics[width=1in,height=1.25in,clip,keepaspectratio]{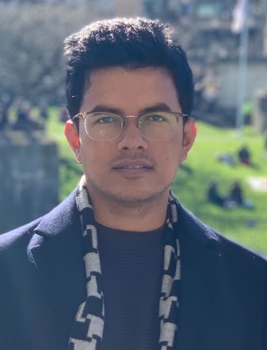}}]{Mobarakol Islam} 
received his Ph.D. degree from the NUS Graduate School for Integrative Sciences and Engineering (NGS), National University of Singapore, in Dec 2019. He is a Postdoctoral Fellow at the Department of Computing, Imperial College London, under Dr. Ben Glocker in BioMedIA Lab. His research is mainly in the interdisciplinary field of deep learning, medical image analysis, and robotic visual perception.
\end{IEEEbiography}

\begin{IEEEbiography}[{\includegraphics[width=1in,height=1.25in,clip,keepaspectratio]{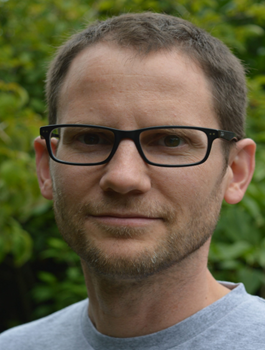}}]{Ben Glocker} 
is Professor in Machine Learning for Imaging at the Department of Computing at Imperial College London where he co-leads the Biomedical Image Analysis Group. He also leads the HeartFlow-Imperial Research Team and is Head of ML Research at Kheiron Medical Technologies. He held a Ph.D. from TU Munich and was a postdoc at Microsoft and a Research Fellow at the University of Cambridge. His research is at the intersection of medical imaging and artificial intelligence, aiming to build safe and ethical computational tools for improving image-based detection and diagnosis of disease.
\end{IEEEbiography}


\begin{IEEEbiography}[{\includegraphics[width=1in,height=1.25in,clip,keepaspectratio]{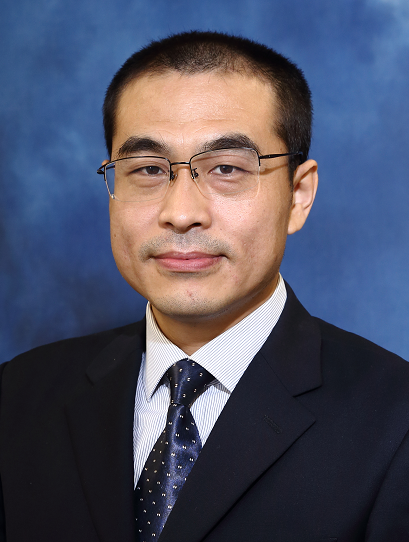}}]{Hongliang Ren}
received his Ph.D. in Electronic Engineering (Specialized in Biomedical Engineering) from The Chinese University of Hong Kong (CUHK) in 2008. He has been navigating his academic journey through the Chinese University of Hong Kong, UC Berkeley, Johns Hopkins University, Children’s Hospital Boston, Harvard Medical School, Children’s National Medical Center, United States, and the National University of Singapore. He serves as an Associate Editor for IEEE Transactions on Automation Science \& Engineering (T-ASE) and Medical \& Biological Engineering \& Computing (MBEC). He has served as an active organizer and contributor on the committees of numerous robotics conferences, including a variety of roles in the flagship IEEE Conf. on Robotics and Automation (ICRA), IEEE Conf. on Intelligent Robots and Systems (IROS), as well as other domain conferences such as ROBIO/BIOROB/ICIA. He served as publicity chair for ICRA 2017, concurrently as Organizing Chair for ICRA 2017 workshop on Surgical Robots, and as video chair for ICRA 2021. He has delivered numerous invited keynotes/talks at flagship conferences/workshops at ICRA/IROS/ROBIO/ICIA. He is the recipient of the IFMBE/IAMBE Early Career Award 2018, the Interstellar Early Career Investigator Award 2018, and the ICBHI Young Investigator Award 2019. He also receives numerous international conference awards, including Best Conference Paper Awards at IEEE ROBIO 2019, IEEE RCAR 2016, IEEE CCECE 2015, IEEE Cyber 2014, and IEEE ROBIO 2013. His research is mainly on Biorobotics \& intelligent systems, medical mechatronics, continuum, and soft flexible robots and sensors, multisensory perception, learning and control in image-guided procedures, deployable motion generation, compliance modulation/sensing, cooperative and context-aware sensors/actuators in human environments, robotic surgery, flexible robotics, and machine artificial intelligence.
\end{IEEEbiography}

\end{document}


\setcounter{page}{1}
\twocolumn[
\centering
\Large
\textbf{Confidence-Aware Paced-Curriculum Learning by Label Smoothing \\for Surgical Scene Understanding} \\
\vspace{0.5em}Supplementary Material \\
\vspace{1.0em}]
\appendix

In this appendix, we include some supplementary material and more experimental results for the proposed P-CBLS approach.

\subsection{Robot Vision in Surgical Scene Understanding}
Surgical scene understanding is crucial for image-guided robot-assisted surgery, and it enables the system to supervise and optimize the surgical procedures, offer context-aware decision support and provide prompt warning of deviant behavior and abnormal operations~\cite{huaulme2020offline}. Tool presence detection is modeled as an multi-label image classification problem ~\cite{twinanda2016endonet}. STGCN~\cite{wang2019graph} implements tool detection by training graph convolutional network with the video segments which contain the labeled frame and the adjacent frames. TMRNet~\cite{jin2021temporal} is proposed to achieve automatic surgical workflow recognition which is a key component for context-aware computer-assisted systems. Automatic surgical report generation ~\cite{xu2021class, xu2021learning} can reduce the workload of document entry task for clinicians. Automatic Instrument Segmentation~\cite{shvets2018automatic, gonzalez2020isinet} is also significant to design intelligent systems in robot-assisted surgery. In this work, we develop a robust curriculum based learning strategy for safety-critical robot vision tasks like robot-assisted surgery.

\begin{table*}[b]
\centering
\caption{\textbf{Tool Segmentation}. IoU per class, mean IoU and mean Dice on Tool Segmentation dataset for DNNs trained normally, DNNs trained using ULS with smoothing factor of $0.1$, and DNNs trained using our proposed CBLS and P-CBLS approach. We show significant improvements using LinkNet34~\cite{chaurasia2017linknet}.}
\scalebox{1}{
\begin{tabular}{cc|ccccccc|c|c}
\toprule
\multicolumn{2}{c|}{\multirow{3}{*}{Model}}                        & \multicolumn{7}{c|}{IoU per tool class}                                                                                                                                                                                                                                                                                                                                                                                                                                                                                                                                                                                                                             & \multirow{3}{*}{\begin{tabular}[c]{@{}c@{}}Mean\\      IoU\end{tabular}} & \multirow{3}{*}{\begin{tabular}[c]{@{}c@{}}Mean\\      Dice\end{tabular}} \\ \cline{3-9}
\multicolumn{2}{c|}{}                                              & \multicolumn{1}{c|}{\multirow{2}{*}{\begin{tabular}[c]{@{}c@{}}Bipolar   \\      Forceps\end{tabular}}} & \multicolumn{1}{c|}{\multirow{2}{*}{\begin{tabular}[c]{@{}c@{}}Prograsp\\      Forceps\end{tabular}}} & \multicolumn{1}{c|}{\multirow{2}{*}{\begin{tabular}[c]{@{}c@{}}Needle\\      Driver\end{tabular}}} & \multicolumn{1}{c|}{\multirow{2}{*}{\begin{tabular}[c]{@{}c@{}}Curved \\      Scissors\end{tabular}}} & \multicolumn{1}{c|}{\multirow{2}{*}{\begin{tabular}[c]{@{}c@{}}Ultrasound\\      Probe\end{tabular}}} & \multicolumn{1}{c|}{\multirow{2}{*}{Suction}} & \multirow{2}{*}{\begin{tabular}[c]{@{}c@{}}Clip\\      Applier\end{tabular}} &                                                                          &                                                                           \\
\multicolumn{2}{c|}{}                                              & \multicolumn{1}{c|}{}                                                                                   & \multicolumn{1}{c|}{}                                                                                 & \multicolumn{1}{c|}{}                                                                              & \multicolumn{1}{c|}{}                                                                                 & \multicolumn{1}{c|}{}                                                                                 & \multicolumn{1}{c|}{}                         &                                                                              &                                                                          &                                                                           \\ \midrule
\multicolumn{1}{c|}{\multirow{2}{*}{Baseline}} & LinkNet34         & \multicolumn{1}{c|}{86.09}                                                                              & \multicolumn{1}{c|}{35.71}                                                                            & \multicolumn{1}{c|}{40.14}                                                                         & \multicolumn{1}{c|}{98.84}                                                                            & \multicolumn{1}{c|}{4.19}                                                                             & \multicolumn{1}{c|}{38.18}                    & 5.13                                                                         & 44.04                                                                    & 53.56                                                                     \\ \cline{2-11} 
\multicolumn{1}{c|}{}                          & LS                & \multicolumn{1}{c|}{89.82}                                                                              & \multicolumn{1}{c|}{28.52}                                                                            & \multicolumn{1}{c|}{49.88}                                                                         & \multicolumn{1}{c|}{97.62}                                                                            & \multicolumn{1}{c|}{1.15}                                                                             & \multicolumn{1}{c|}{32.20}                    & 0.05                                                                         & 42.75                                                                    & 50.78                                                                     \\ \hline
\multicolumn{1}{c|}{\multirow{6}{*}{Ours}}     & CBLS (ULS)        & \multicolumn{1}{c|}{91.61}                                                                              & \multicolumn{1}{c|}{36.20}                                                                            & \multicolumn{1}{c|}{65.68}                                                                         & \multicolumn{1}{c|}{98.51}                                                                            & \multicolumn{1}{c|}{0.80}                                                                             & \multicolumn{1}{c|}{38.66}                    & 0.00                                                                         & 47.35                                                                    & 54.95                                                                     \\ \cline{2-11} 
\multicolumn{1}{c|}{}                          & CBLS (SVLS)       & \multicolumn{1}{c|}{83.48}                                                                              & \multicolumn{1}{c|}{25.99}                                                                            & \multicolumn{1}{c|}{28.08}                                                                         & \multicolumn{1}{c|}{\textbf{98.73}}                                                                   & \multicolumn{1}{c|}{5.83}                                                                             & \multicolumn{1}{c|}{42.93}                    & \textbf{57.34}                                                               & 48.91                                                                    & 59.92                                                                     \\ \cline{2-11} 
\multicolumn{1}{c|}{}                          & CBLS (ULS+SVLS)   & \multicolumn{1}{c|}{88.61}                                                                              & \multicolumn{1}{c|}{43.16}                                                                            & \multicolumn{1}{c|}{57.48}                                                                         & \multicolumn{1}{c|}{98.56}                                                                            & \multicolumn{1}{c|}{5.47}                                                                             & \multicolumn{1}{c|}{42.45}                    & 18.66                                                                        & 50.63                                                                    & 61.14                                                                     \\ \cline{2-11} 
\multicolumn{1}{c|}{}                          & P-CBLS (ULS)      & \multicolumn{1}{c|}{\textbf{91.69}}                                                                     & \multicolumn{1}{c|}{31.33}                                                                            & \multicolumn{1}{c|}{\textbf{67.13}}                                                                & \multicolumn{1}{c|}{98.22}                                                                            & \multicolumn{1}{c|}{\textbf{10.32}}                                                                   & \multicolumn{1}{c|}{40.42}                    & 1.88                                                                         & 48.71                                                                    & 57.54                                                                     \\ \cline{2-11} 
\multicolumn{1}{c|}{}                          & P-CBLS (SVLS)     & \multicolumn{1}{c|}{89.20}                                                                              & \multicolumn{1}{c|}{39.85}                                                                            & \multicolumn{1}{c|}{40.69}                                                                         & \multicolumn{1}{c|}{97.73}                                                                            & \multicolumn{1}{c|}{9.52}                                                                             & \multicolumn{1}{c|}{51.83}                    & 43.79                                                                        & 53.23                                                                    & 64.94                                                                     \\ \cline{2-11} 
\multicolumn{1}{c|}{}                          & P-CBLS (ULS+SVLS) & \multicolumn{1}{c|}{85.42}                                                                              & \multicolumn{1}{c|}{\textbf{59.37}}                                                                   & \multicolumn{1}{c|}{32.31}                                                                         & \multicolumn{1}{c|}{97.71}                                                                            & \multicolumn{1}{c|}{4.57}                                                                             & \multicolumn{1}{c|}{\textbf{52.73}}           & 50.87                                                                        & \textbf{54.71}                                                           & \textbf{65.65}                                                            \\ \bottomrule
\end{tabular}}
\label{table:segmentation_class_wise}
\end{table*}

\subsection{Segmentation results}
\begin{figure*}[!hbpt]
\centering
\includegraphics[width=1\linewidth]{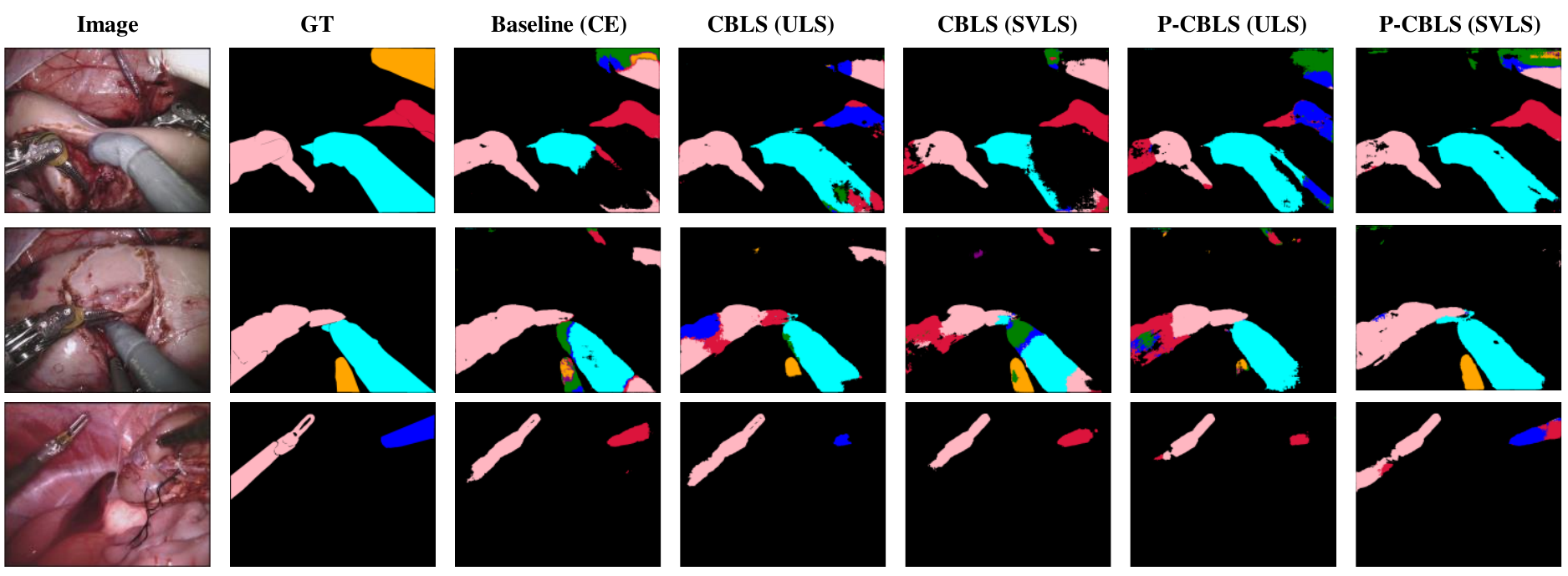}
\caption{\textbf{Visualization of the predicted mask of LinkNet34~\cite{chaurasia2017linknet} model based on different proposed approaches}. Bipolar forceps, prograsp forcep, monopolar curved scissor, and suction are indicated in pink, red, blue, and yellow. P-CBLS here specifically refers to pixel wise paced learning}
\label{fig:segmentation_result_supp}
\end{figure*}

We show the class wise segmentation results based on LinkNet34 in Table~\ref{table:segmentation_class_wise}. Our proposed methods show surprising improvement in terms of the ``Clip Applier" class. We also visualize the predicted masks from the baseline and our approaches based on LinkNet34~\cite{chaurasia2017linknet} in Figure~\ref{fig:segmentation_result_supp}.

\begin{figure*}[!hbpt]
\centering
\includegraphics[width=1\linewidth]{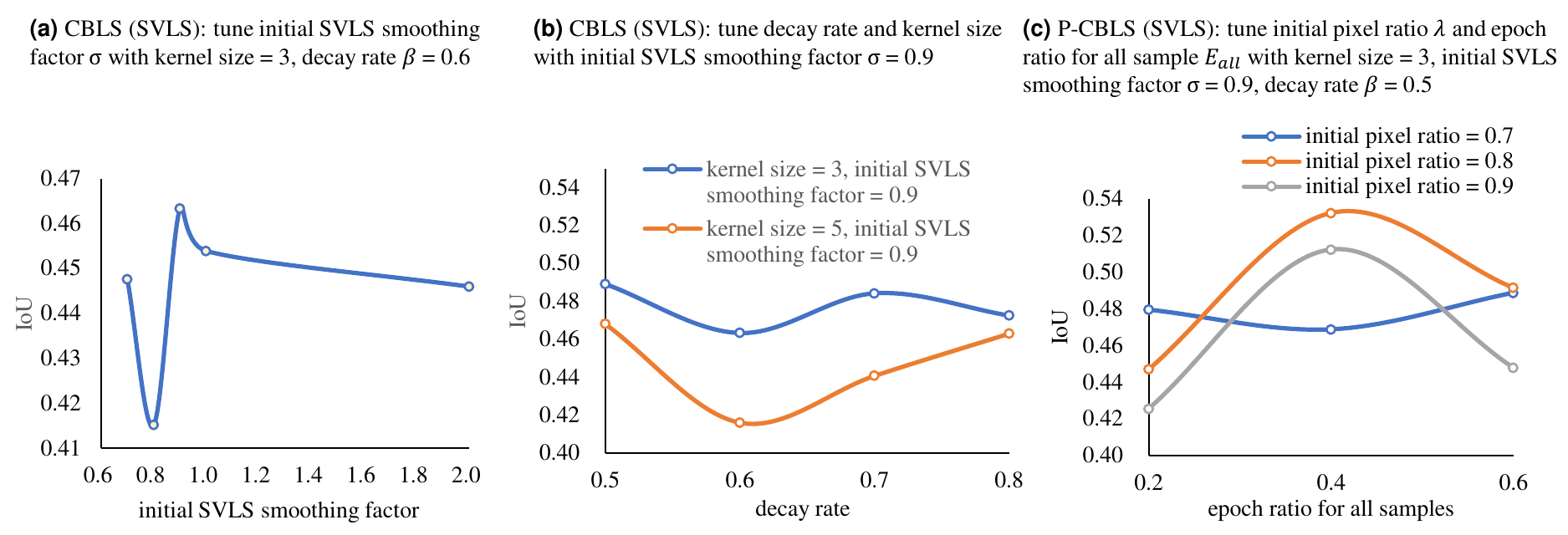}
\caption{\textbf{Ablation study based on hyperparameters of our CBLS and P-CBLS on Tool Segmentation dataset with LinkNet34 model.} (a) and (b) hyperparameters of CBLS (SVLS). (c) hyperparameters of P-CBLS (SVLS).}
\label{fig:ablation_IoU}
\end{figure*}

\vspace{40pt}

\subsection{Ablation study}
\subsubsection{\textbf{Hyperparameters of P-CBLS (ULS)}}
The ablation study on Surgical Captioning dataset is shown in Table \ref{table:ablation_on_surgical_captioning}. We report the captioning performance under different ULS smoothing factors with the decay rate of $0.95$ and the calibration performance with the Brier.

\subsubsection{\textbf{Hyperparameters of pixel wise P-CBLS (SVLS)}}
We analyze the effect of different initial SVLS smoothing factor $\sigma$, decay rate $\beta$, kernel size $k$, initial pixel ratio $\lambda$ and epoch ratio for all samples $E_{all}$ for LinkNet34~\cite{chaurasia2017linknet} on Tool Segmentation dataset. In CBLS (SVLS) experimetns, we tune the initial SVLS smoothing factor $\sigma$ = [0.7, 0.8, 0.9, 1.0, 2.0] with the decay rate $\beta$ of 0.6 and kernel size $k$ = $3$, as shown in Figure~\ref{fig:ablation_IoU} (a). ``$k$ = 3, $\sigma$ = 0.9, $\beta$ = 0.6'' provides the best IoU. In Figure~\ref{fig:ablation_IoU} (b), we study the effect of decay rate $\beta$ and kernel size $k$. In specific, we tune the decay rate $\beta$ with range $\beta \in [0.5, 0.8]$ at a step of $0.1$ and the kernel size $k$ = [3, 5] with the initial SVLS smoothing factor $\sigma$ = 0.9. ``$k$ = 3, $\sigma$ = 0.9, $\beta$ = 0.5'' provides the best IoU. In pixel wise P-CBLS (SVLS) experiments, we tune the initial pixel ratio $\lambda$ with range $\lambda \in [0.7, 0.9]$ at a step of $0.1$ and the epoch ratio for all samples $E_{all}$ with range $E_{all} \in [0.2, 0.6]$ at a step of $0.2$ with the ``$k$ = 3, $\sigma$ = 0.9, $\beta$ = 0.5'' which provides the best IoU in previous CBLS (SVLS) experimetns. The initial pixel ratio $\lambda$ of $0.8$, epoch ratio for all samples $E_{all}$ of $0.4$ generates the best IoU, as shown in Figure~\ref{fig:ablation_IoU} (c).

\begin{table}[H]
\centering
\caption{\textbf{Ablation study based on different smoothing on Surgical Captioning dataset.} We report the captioning performance with METEOR and CIDEr and calibration performance with the commonly used Brier.}
\scalebox{1}{
\begin{tabular}{c|c|c|c|c}
\hline
\multirow{2}{*}{\begin{tabular}[c]{@{}c@{}}smoothing\\      factor\end{tabular}} & \multirow{2}{*}{\begin{tabular}[c]{@{}c@{}}decay \\      rate\end{tabular}} & \multirow{2}{*}{METEOR} & \multirow{2}{*}{CIDEr} & \multirow{2}{*}{Brier$\downarrow$} \\
                                                                                 &                                                                               &                         &                        &                        \\ \hline
0.06                                                                             & \multirow{3}{*}{0.95}                                                         & 0.3398                  & 2.9973                 & 0.6488                 \\ \cline{1-1} \cline{3-5} 
0.08                                                                             &                                                                               & 0.3423                  & 2.9713                 & 0.6202                 \\ \cline{1-1} \cline{3-5} 
0.10                                                                             &                                                                               & 0.3652                  & 3.0965                 & 0.5990                 \\ \hline
\end{tabular}}
\label{table:ablation_on_surgical_captioning}
\end{table}

\newpage
\bibliographystyle{IEEEtran}
\bibliography{sample}